\documentclass[11pt]{article}
\usepackage{coling2020}
\usepackage{times}
\usepackage{url}
\usepackage{latexsym}
\usepackage{todonotes}
\usepackage{booktabs}
\usepackage{multirow}
\usepackage{amsmath}
\usepackage{bm}
\usepackage[colorlinks=true,
            linkcolor=red,
            urlcolor=blue,
            citecolor=blue]{hyperref}
\usepackage{caption}
\usepackage{subcaption}
\usepackage{soul}
\usepackage{enumitem}
\usepackage{graphicx}
\usepackage{tablefootnote}
\usepackage{amssymb}
\colingfinalcopy % Uncomment this line for the final submission

\title{Recent Neural Methods on Slot Filling and Intent Classification \\for Task-Oriented Dialogue Systems: A Survey}

\author{Samuel Louvan \\
  University of Trento \\
  Fondazione Bruno Kessler \\
  {\tt slouvan@fbk.eu} \\\And
  Bernardo Magnini \\
  Fondazione Bruno Kessler \\
  {\tt magnini@fbk.eu} \\}

\date{}

\begin{document}
\maketitle
\begin{abstract}
% pertama harus ngomongin perkembangan yang menarik di area dialgoue systems terus 
% SLU itu penting
% terus paper ini ngapain
% harapannya apa dengan paper ini
In recent years, fostered by deep learning technologies and by the high demand for conversational AI, various approaches have been proposed  that address the capacity to elicit and understand user’s needs in task-oriented dialogue systems. We focus on two core tasks,   slot filling (SF) and intent classification (IC), and survey how neural based models have rapidly evolved to address natural language understanding in dialogue systems. We introduce three neural architectures: \textit{independent models}, which model SF and IC separately, \textit{joint models},  which exploit the mutual benefit of the two tasks simultaneously, and \textit{transfer learning models},  that scale the model to new domains.  We  discuss the current state of the research in SF and IC, and highlight challenges that still require attention.

\end{abstract}

\section{Introduction}
\blfootnote{
    \hspace{-0.65cm}  % space normally used by the marker
    This work is licensed under a Creative Commons 
    Attribution 4.0 International License.
    License details:
    \url{http://creativecommons.org/licenses/by/4.0/}.
}
The ability to understand  user's requests is essential to develop effective task-oriented dialogue systems.  For example, in the utterance "\textit{I want to listen to Hey Jude by The Beatles}", a dialogue system should correctly identify that the user's intention is to give a command  to play a song, and that \textit{Hey Jude} and \textit{The Beatles} are, respectively, the song's title  and the artist name that the user would like to listen. In a dialogue system this information is typically represented through a \textit{semantic-frame} structure \cite{tur2011spoken}, 
%as shown in Table \ref{tab:semantic_frame_example}.
and extracting such representation involves two tasks: identifying the correct frame (i.e.  \textit{intent classification (IC)}) and filling the correct value for the slots of the frame (i.e.  \textit{slot filling (SF)}). 

In recent years, neural-network based models have achieved the state of the art  for a wide range of natural language processing tasks, including SF and IC. Various neural architectures have been experimented on SF and IC, including RNN-based  \cite{Mesnil2013InvestigationOR} and attention-based \cite{Liu2016AttentionBasedRN} approaches, till the more recent transformers models \cite{Chen2019BERTFJ}.  Input representations have also evolved from  static word embeddings \cite{Mikolov2010RecurrentNN,Collobert2008AUA,Pennington2014GloveGV} to contextualized word embeddings \cite{peters2018deep,Devlin2019BERTPO}.  Such progress allows to better address dialogue phenomena involving SF and IC, including  context modeling, handling out-of-vocabulary words, long-distance dependency between words, and to better exploit the  synergy between SF and IC through joint models. 
In addition to rapid progresses in the research community, the demand for commercial conversational AI is also growing fast, shown by a variety of available solutions, such as Microsoft LUIS, Google Dialogflow, RASA, and Amazon Alexa. These solutions also use various kinds of semantic frame representations as part of their framework.

Motivated by the rapid explosion of scientific progress, and by unprecedented market attention,  we think that a guided map of the approaches on SF and IC  can be useful for a large spectrum of researchers and practitioners interested in dialogue systems.
The primary goal of the  survey is to give a broad overview of  recent neural models applied to SF and IC, and to compare their performance in the context of task-oriented dialogue systems.  We also highlight and discuss open issues that still need to be addressed in the future. The paper is structured as follows: Section \ref{sec:slot_filling_intent_classification} describes the SF and IC tasks,   commonly used datasets and evaluation metrics. Section \ref{sec:independent_model}, \ref{sec:joint}, and \ref{sec:transfer_learning} elaborate on the progress and state of the art of \textit{independent}, \textit{joint}, and \textit{transfer learning models }for both tasks. Section \ref{sec:discussion} discusses the performance of existing models and   open challenges.

% \footnote{https://www.luis.ai/home}
% \footnote{https://dialogflow.com/}
% \footnote{https://rasa.com/docs/rasa/}
% \footnote{https://developer.amazon.com/en-US/docs/alexa/custom-skills/create-intents-utterances-and-slots.html}

\begin{table*}[ht!]
    \centering
    \small
    \begin{tabular}{|l||c|c|c|c|c|c|c|c|c|c|c|}
    
    % \multicolumn{11}{|c|}{\textbf{Semantic Frame}} \\
    \hline
    \textbf{Utterance} & I & want & to & listen  & to & Hey & Jude & by  & The & Beatles       \\
    \hline
    \textbf{Slot}   & \textsc{O}  & \textsc{O}       & \textsc{O}    & \textsc{O}& \textsc{O}  & \textsc{B-SONG} & \textsc{I-SONG}    & \textsc{O} & \textsc{B-ARTIST}  & \textsc{I-ARTIST}\\
    \hline
    \multicolumn{1}{|l||}{\textbf{Intent}} & \multicolumn{10}{|l|}{\textsc{play\_song}}  \\
    \hline
    \end{tabular}
    \caption{Example of SF and IC output for an utterance. Slot labels are in \textsc{BIO} format: \textsc{B} indicates the start of a slot span, \textsc{I}  the inside of a span while \textsc{O} denotes that the word does not belong to any slot. }
    \label{tab:sf_ic_example}
\end{table*}

% \todo[inline]{Explain the structure of the paper}
\section{Slot Filling and Intent Classification}
\label{sec:slot_filling_intent_classification}
This section provides some background relevant for SF and IC, sets the scope of the survey with respect to the context in dialogue systems, defines SF and IC as tasks, and introduces the datasets and the metrics that will be used in the rest of the paper.

\subsection{Background}
\label{subsec:background}
Task-oriented dialogue systems aim to assist users to accomplish a task (e.g. booking a flight, making a restaurant reservation and playing a song) through dialogue in natural language, either in a spoken or written form.
%a ,  consisting of one or more turns of user's utterances and the system's responses. 
%Going from the user utterance as input to the system's output response involves  a pipeline of components.
As in most of the current approaches, we assume a system involving a pipeline of components \cite{young2010hidden}, where the user utterance is first processed by an Automatic Speech Recognition (ASR) module and then
processed by a Natural Language Understanding (NLU) component, which interprets the user's needs. Then a Dialogue State Tracker (DST) accumulates the dialogue information as the conversation progresses and may query a domain knowledge base to obtain relevant data. A dialogue policy manager then decides what is the next action to be executed and a Natural Language Generation (NLG) component produces the actual response to the user. 

We focus on the NLU component, and we generalize several recent approaches assuming that the output of the NLU process is a partially filled semantic frame \cite{wang2005spoken,tur2011spoken}, corresponding to the intent of the user in a certain portion of the dialogue, with a number of slot-value pairs that need to be filled to accomplish the intent.
The notion of \textit{intent} originates from the idea that utterances can be assigned to a small set of \textit{dialogue acts} \cite{stolcke-etal-2000-dialogue}, and it is now largely adopted to identify a task or action that the system can execute in a certain domain. \textit{Slot-value pairs}, on the other end, represent the  domain of the dialogue, and have been actually implemented either as an ontology \cite{conf/interspeech/Bellegarda13}, possibly with reasoning services (e.g. checking the constraints over slot values) or simply trough a list of entity types that the system needs to identify during the dialogue.

Intents may correspond either to specific needs of the user (e.g. blocking a credit card, transferring money, etc.), or to general needs (e.g. asking for clarification, thanking, etc.). Slots are defined for each intent: for instance, to block a credit card it is relevant to know the name of the owner and the number of the card. Values for the slots are collected through the dialogue, and can be expressed by the user either in a single turn or in several turns. 
At each user turn in the dialogue the NLU component has to determine the intent of the user utterance (\textit{intent classification}) and has to detect the slot-value pairs referred in the particular turn (\textit{slot filling}).
Table \ref{tab:sf_ic_example} shows the expected NLU output for the  utterance "\textit{I want to listen to Hey Jude by The Beatles}".

\subsection{Scope of the Survey}

In Section \textsection\ref{subsec:background}, we described a task-oriented system as a pipeline of components, saying that SF and IC are core tasks at the NLU level.  Particularly, IC consists of classifying an utterance with a set of pre-defined intents, while SF is defined as a sequence tagging problem \cite{Raymond2007GenerativeAD,Mesnil2013InvestigationOR}, where each token of the utterance has to be tagged with a slot label. In this scenario training data for SF typically consist of single utterances in a dialogue where tokens are annotated with a pre-defined set of slot names, and slot values correspond to arbitrary sequences of tokens. In this perspective, it is worth mentioning a research line on dialogue state tracking \cite{henderson2014word,mrkvsic2015multi,budzianowski-etal-2018-multiwoz}, where the NLU component is usually embedded into DST.  What is relevant for our topic is that in this context SF is defined as a classification problem: given the current utterance and the previous dialogue history, the system has to decide whether a certain slot-value pair defined in the domain ontology is referred or not in the current utterance. Although promising, from the NLU perspective, this research line poses constraints (e.g. all slot-value pairs have to be pre-defined in an ontology,) that limit the SF applicability. For this reason, and because NLU components are the prevalent solution in current task-oriented systems, the focus of our survey will be on SF as a sequence tagging problem, as more precisely defined in the next section.

\subsection{Task Definition}

\label{sec:approach}
We formulate SF and IC as follows. 
Given an input utterance $\boldsymbol{x} = (x_1,x_2,.., x_T)$, SF consists in a token-level sequence tagging, where the system has to assign a corresponding slot label $\boldsymbol{y}^{slot} = (y^{slot}_1, y^{slot}_2,.., y^{slot}_T)$ to each token $x_i$ of the utterance. 
On the other end, IC is defined as a classification task over utterances, where the system has to assign the correct intent label $y^{intent}$ for the whole utterance $\boldsymbol{x}$. In general, most machine learning approaches learn a probabilistic model to estimate $p(y^{intent},\boldsymbol{y}^{slot}|\boldsymbol{x}, \boldsymbol{\theta})$ where $\boldsymbol{\theta}$ is the parameter of the model. Table \ref{tab:sf_ic_example} shows an example of the expected output of a model for the SF and IC tasks. In the following sections, we outline the main models that have been proposed for SF and IC, and  categorize the models into three groups, namely \textit{independent models} (\textsection \ref{sec:independent_model}) , \textit{joint models} (\textsection \ref{sec:joint}), and \textit{transfer learning based models} (\textsection{\ref{sec:transfer_learning}}). 
% These categories are not strictly mutual exclusive as joint model can be used for independent model but not the other way around. 

\subsection{Datasets for SF and IC}
\label{dataset}
%As in other NLP areas, the availability of training data has influenced both the definition of the tasks and of the approaches. 
% We split datasets for IC and SF in two groups: single-turn datasets and multiple-turn datasets. Overall statistics  are reported in Table~\ref{dataset-table}.  

In this section, according to our task definition, we list available dialogue datasets (most of them are publicly available) where each utterance is assigned to one intent, and tokens are annotated with slot names. Most of such datasets are collections of \textit{single turn user utterances} (i.e., not multi-turn dialogues). An example of a single-turn utterance annotation is shown in Table \ref{tab:sf_ic_example}. 

The ATIS (Airline Travel Information System) dataset \cite{Hemphill1990TheAS} is the most widely used single-turn dataset for NLU benchmarking. The total number of utterances is around 5K utterances that consist of queries related to the airline travel domain, such as searching for a flight, asking for flight fare, etc. While it has a relatively large number slot and intent labels, the distribution is quite skewed; more than 70\% of the intent is a flight search. The slots are dominated by a slot that expresses location names such as \textsc{FromLocation} and \textsc{ToLocation}. The MEDIA dataset \cite{DBLP:conf/interspeech/Bonneau-MaynardRAKM05} is constructed by simulating the conversation between a tourist and a hotel representative in the French language. Compared to ATIS, the MEDIA corpus size is around three times larger; however, MEDIA is only annotated with slot labels. The slots are related to hotel booking scenarios such as the number of people, date, hotel facility, relative distance, etc. The MIT corpus \cite{DBLP:conf/icassp/LiuPCG13} is constructed through a crowdsourcing platform where crowd workers are hired to create natural language queries in English and annotate the slot label in the queries. The MIT corpus covers two domains, namely movie and restaurant, in which the utterances are related to finding information of a particular movie or actor, searching or booking a restaurant with a particular distance and cuisine criteria. The SNIPS dataset \cite{Coucke2018SnipsVP} was collected by  crowdsourcing through the SNIPS voice platform. Intents include requests to a digital assistant to complete various tasks, such as asking the weather, playing a song, book a restaurant, asking for a movie schedule, etc. SNIPS is now often used as a benchmark for NLU evaluations.

While most datasets are available in English, recently there has been growing interest in expanding slot filling and intent classification datasets to non-English languages. The original ATIS dataset has been derived into several languages, namely Hindi, Turkish \cite{Upadhyay2018AlmostZC}, and Indonesian \cite{Susanto2017NeuralAF}. The MultiATIS++ dataset from \newcite{DBLP:journals/corr/abs-2004-14353} expands the ATIS dataset to more languages, namely  Spanish, Portuguese, German, French, Chinese, and Japanese. The work from \cite{DBLP:conf/clic-it/BellomariaCFR19} introduces the Italian version of the original SNIPS dataset. The Facebook multi-lingual dataset \cite{Schuster2018CrosslingualTL},  introduced a dataset on Thai and Spanish languages  across three domains namely weather, alarm, and reminder. 
The detailed statistics of each dataset are listed in Appendix A.

\subsection{Evaluation Metrics}
 For the IC task, evaluation is performed on the utterance level. The typical evaluation metric for IC is \textit{accuracy},  calculated as the number of the correct predictions made by the model divided by the total number of predictions. As for SF, the evaluation is performed on the entity level. The common metrics used is the metric introduced in CoNLL-2003 shared task  \cite{DBLP:conf/conll/SangM03} to evaluate Named Entity Recognition (NER) by computing the F-1 score. The F1-score, is the harmonic mean score between precision and recall. Precision  is the percentage of slot predictions from the model which are correct, while  recall  is the percentage of slots in the corpus that are found by the model. A slot prediction is considered \textit{correct} when an \textit{exact} match is found \cite{DBLP:conf/conll/SangM03}. As the slot is annotated in BIO format to mark the boundary of the slot (see Table \ref{tab:sf_ic_example}), a correct prediction is only counted when the model can predict the correct slot label on the correct token offset. Consequently, the exact match metrics does not reward cases when the model predict correct slot label but get the incorrect slot boundary (\textit{partial match}).  

\section{Independent Models for SF and IC}
\label{sec:independent_model}
% The original problem of $p(y^{intent},\textbf{y}^{slot}|\textbf{x}, \boldsymbol{\theta})$ is further factorized  into $p(y^{intent}|\textbf{x}, \boldsymbol{\theta^{slot}})$  and $p(\boldsymbol{y}^{slot}|\textbf{x}, \boldsymbol{\theta^{intent}})$, where $\boldsymbol{\theta^{slot}}$ and $\boldsymbol{\theta^{intent}}$ are the learned parameters for the SF and IC models, respectively. 
Independent models train each task \textit{separately} and recent neural models typically use RNN as the building block for SF and IC. At each time step \textit{t}, the encoder transforms the word representation $x_t$ to the hidden state $h_t$. For SF, the output layer predicts the slot label $y^{slot}_t$  condition on $h_t$. For IC, typically the last hidden state $h_T$ is used to predict the intent label $y^{intent}$of the utterance $\boldsymbol{x}$. Note that, for independent approaches, the models for SF and IC are trained separately.  Most neural models for SF and IC generally consist of several layers, namely an \textit{input layer}, one or more \textit{encoder layer}, and an \textit{output layer}. Consequently, the main differences between models are in the specifics of these layers. The most common dataset used for evaluating independent models is  ATIS.

In the \textit{input layer} of neural models each word is mapped into embeddings.  \newcite{Mesnil2013InvestigationOR} compared several embeddings, namely pre-trained SENNA \cite{Collobert2011NaturalLP}, RNN Language Model (RNNLM) \cite{Mikolov2011RNNLMR}, and random embeddings. SENNA gives the best result compared to other embeddings, and, typically, further fine-tuning  word embeddings  improves performance. \cite{Yao2013RecurrentNN} report that  embeddings learned from scratch directly on ATIS data (\textit{task-specific embeddings}) are better than SENNA. However,  task-specific embeddings are composed not only by words but also by named entities (\textit{NE}) and syntactic features\footnote{Gold named entity and syntactic information}. NE improves performance significantly while part-of-speech only adds small benefits. \newcite{Ravuri2015RecurrentNN} emphasizes the importance of \textit{character representation} to handle OOV issues. 
% They demonstrate that a single \textit{multi-task} deep neural network model can achieve good performance for several NLP tasks while only using continuous word representation (\textit{embedding}) as input. 

For the \textit{encoder layer}, various RNN architectures have been applied to SF and IC \cite{Mesnil2013InvestigationOR,Mesnil2015UsingRN,Liu2015RecurrentNN}. \newcite{Mesnil2013InvestigationOR} compare the Elman \cite{elman1990finding} and Jordan \cite{jordan1997serial} RNNs. They observe that the performance of the Jordan RNN is marginally better than Elman. They also experiment  a \textit{bi-directional} version of Jordan RNN and obtained the best score of 93.89 F1 for SF, performing better than CRF for about +1 absolute F1 improvement. \newcite{xu2013convolutional} use Convolutional Neural Network (CNN) \cite{LeCun1998GradientbasedLA} to extract 5-gram features and apply max-pooling to obtain the word representation before passing it to the output layer. Compared with RNN \cite{Yao2013RecurrentNN,Mesnil2013InvestigationOR},  CNN gives lower performance for SF on ATIS. Other studies \cite{Yao2014SpokenLU,vu2016bi} adapt Long Short-Term Memory Network (LSTM) \cite{Hochreiter1997LongSM} to SF. The LSTM model gives better SF performance compared to CRF, CNN, and RNN. \newcite{Ravuri2015RecurrentNN} compare the performance of vanilla RNN and LSTM for IC. They find that the vanilla RNN works best for shorter utterances, while LSTM is better for longer utterances. 

% \citet{Yao2014SpokenLU} do not use pre-trained word embedding and learn  embeddings (3-grams) from scratch. 

\begin{table}[t!]
\small
\begin{center}
\begin{tabular}{llllll}
\toprule

\bf   & \bf Input  & \textbf{Model (Enc/Dec)} & \textbf{Output}  & \textbf{Slot (F1)} & \textbf{Intent(Err)}\\
\toprule
\newcite{xu2013convolutional} & lexical & CNN & softmax & 94.35 & 6.65\\
\newcite{Yao2013RecurrentNN}  & lexical & Elman RNN & softmax & 94.11 &- \\
\newcite{Yao2013RecurrentNN}  & lexical+NE & Elman RNN & softmax & 96.60 -\\

\newcite{Yao2014SpokenLU}& lexical & LSTM & softmax & 94.85 & -\\
\newcite{Yao2014RecurrentCR}& lexical+NE & Elman RNN & CRF & 96.65 & -\\
\newcite{Mesnil2015UsingRN}&  lexical & Hybrid Elman + Jordan RNN & softmax & 95.06 & -\\
\newcite{Liu2015RecurrentNN}&  lexical & Elman RNN with label sampling & softmax & 94.89 & -\\
\newcite{vu2016bi} & lexical & bi-directional RNN & softmax &94.92&- \\
\newcite{Liu2016AttentionBasedRN} & lexical & bi-directional RNN+attention& softmax & 95.75 & 2.35\\
\newcite{DBLP:conf/emnlp/KurataXZY16} & lexical & Encoder-Decoder LSTM& softmax & 95.40 & -\\
% \cite{Yao2014SpokenLU}&  & LSTM & softmax & 95.08 & -\\
% \cite{Yao2014SpokenLU}&  & LSTM & softmax & 95.08 & -\\
% \cite{Yao2014SpokenLU}&  & LSTM & softmax & 95.08 & -\\
% \cite{Yao2014SpokenLU}&  & LSTM & softmax & 95.08 & -\\
\bottomrule
\end{tabular}
\end{center}
\caption{ Comparison of independent SF and IC models and their performance on ATIS.}
\label{tab:independent_models}
\end{table}

% \vspace{-0.05cm}
For the \textit{output layer}, typically a \textit{softmax} function is used for prediction at a particular time step. \newcite{Yao2014RecurrentCR} propose a \textsc{R-CRF} model  combining the feature learning power of RNN and the \textit{sequence level optimization} of CRF for SF. 
The RNN + CRF scoring mechanism incorporates the features learned from RNN and the transition scores of the slot slot labels. R-CRF outperforms CRF and vanilla RNN on ATIS and on the Bing query understanding dataset. Table \ref{tab:independent_models} summarizes the performance of independent models on SF and IC.

\textbf{Takeaways on independent SF and IC models:}
\begin{itemize}[noitemsep,nolistsep, leftmargin=*]
    \item Performance of  RNN encoders (\textit{unidirectional}) are $\text{Jordan}\leq\text{Elman}<\text{LSTM}$. Bi-directional encoding is additive to the performance of each encoder.  
    \item Incorporating more context information is better for SF performance. Using global context information, such as sentence level representation, and attention mechanisms \cite{DBLP:conf/emnlp/KurataXZY16,Liu2016AttentionBasedRN} boosts performance of bi-directional encoder even further. 
    \item When adding external features is possible, semantic features such as NE are more beneficial than syntactic features for SF. When NE is used, it can boost the model performance for SF significantly.
    \item The slot filling task is related to Named Entity Recognition (NER) \cite{DBLP:conf/coling/GrishmanS96} task as slot values can be a named entity such as airline name, city name etc. If the slot filling task is modeled as a sequence tagging problem, basically  recent neural models proposed for NER can be used for slot filling and vice versa. To know more about the recent development of neural NER models, one can consult the survey from \newcite{yadav-bethard-2018-survey}. 
    \item The main disadvantage of  independent models is that they do not exploit the interaction between intent and slots and may introduce error propagation when they are used in a pipeline. 
    % \item  IC is relatively easier than SF. 
\end{itemize}

\section{Joint Models for SF and IC}
\label{sec:joint}

\begin{figure*}[htb!]
     \centering
     \includegraphics[scale=0.26]{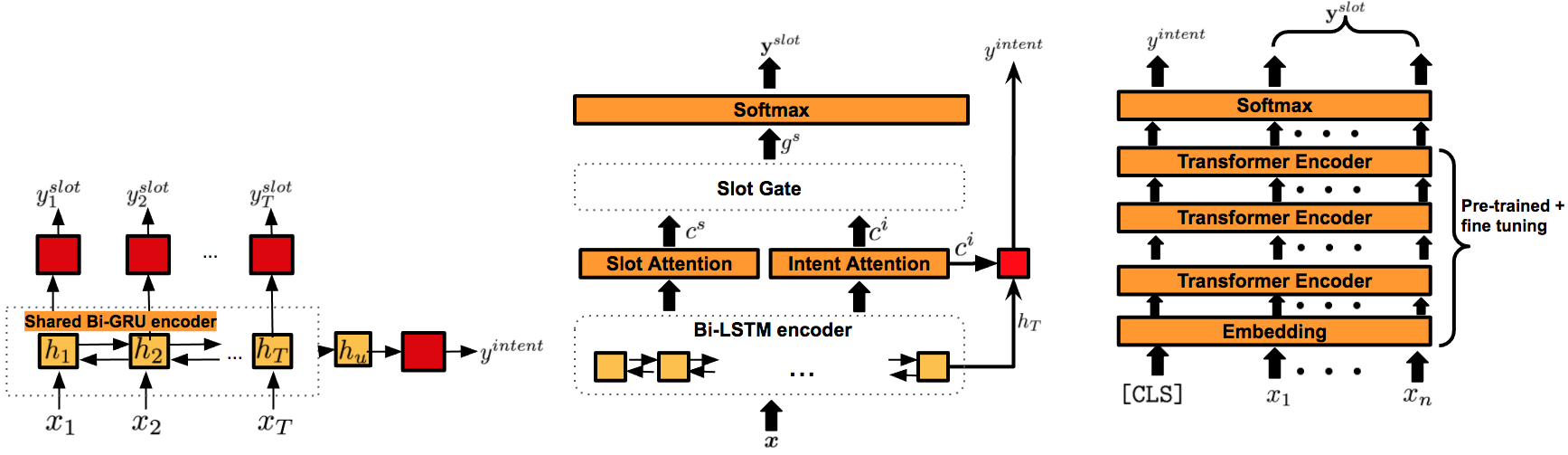}
    \caption{\textit{Left:} Shared Bi-GRU encoder \protect\cite{Zhang2016AJM}. \textit{Middle:} Slot-Gate Mechanism \protect\cite{goo2018slot}. \textit{Right:} BERT Based \protect \cite{Chen2019BERTFJ}.}
    \label{fig:sharing_gating_mechanism}
\end{figure*}

In Section \ref{sec:independent_model} we reported  approaches  that treat SF and IC \textit{independently}. However, as the two tasks always appear together in an utterance and they share information, it is intuitive to think that they can benefit each other. For instance, if the word "\textit{The Beatles}"  is recognized as the slot \textsc{Artist}, then it is more likely that the intent of the utterance is \textsc{PlaySong} rather than \textsc{BookFlight}. On the other hand, recognizing that the intent is \textsc{PlaySong} would help to recognize "\textit{Hey Jude}" as the slot \textsc{Artist} rather than \textsc{MovieName}. 
    
Recent approaches  model the relationship between SF and IC \textit{simultaneously} in a \textit{joint model}.   These approaches promote \textit{two-way} information sharing between the two tasks instead of a one-way (\textit{pipeline}). We describe several alternatives to exploit the relation between SF and IC: through \textit{parameter and state sharing} and \textit{gate mechanism}.

\subsection{Parameter and State Sharing} 
A pioneering work in joint modeling is \newcite{xu2013convolutional}, which performs parameter sharing and captures the relation between SF and IC through Tri-CRF \cite{jeong2008triangular}.  The model uses CNN as a \textit{shared} encoder for both tasks and the produced hidden states are utilized for SF and IC. In addition to features learned from the NN and from the slot label transition,  Tri-CRF incorporates an additional factor $g$ to learn the correlation between the slot label assigned to each word and the intent  assigned to the utterance, which explicitly captures the dependency between the two tasks.  A similar approach  \cite{guo2014joint}, shares the node representation produced by Recursive Neural Network (RecNN) which operates on the syntactic  tree of the utterance. The node's representation is \textit{shared} among SF and IC. \newcite{Zhang2016AJM} use a \textit{shared} bi-GRU encoder and a \textit{joint loss function} between SF and IC (Figure \ref{fig:sharing_gating_mechanism} \textit{Left}), in which the loss function has weights associated with each tasks. 

% Their work obtains substantial improvement on IC with +3\% absolute accuracy increase compared to \citet{guo2014joint}, and marginal improvement of +0.5\% for SF.

\newcite{Liu2016AttentionBasedRN} use  a neural sequence to sequence (encoder-decoder) model with attention mechanism commonly used for neural machine translation. The \textit{shared} encoder is a bi-directional LSTM, and the last hidden state of the encoder is then used by the decoder to generate a sequence of slot labels, while for IC there is a separate decoder. The attention mechanism is used to learn alignments between  slot labels in the decoder and words in the encoder. \newcite{HakkaniTr2016MultiDomainJS} also adopt parameter sharing similar to \newcite{Zhang2016AJM}, but instead of using GRU they use a shared LSTM and perform predictions for slots, intent, and also domain. 

% This approach achieves +2\% F1 absolute improvement over \citet{guo2014joint} for SF. 

 In a  recent approach by \newcite{Wang2018ABB} propose a bi-model based structure to learn the \textit{cross-impact} between SF and IC. They argue that a single model for two tasks can hurt performance, and, instead of sharing  parameters, they use two-task networks to learn the cross-impact between the two tasks and only share the hidden state of the other task. In the model, every hidden state $h^1_t$ in the first network is combined with the hidden state of the second network $h^2_t$, and vice versa. Training is also done asynchronously, as each model has a separate loss function.  Qin et al. \newcite{Qin2019ASF}  use a self-attentive shared encoder to produce better context-aware representations, then apply IC at the \textit{token level} and use this information to guide the SF task. They argue that previous work based on \textit{single utterance-level} intent prediction is more prone to error propagation. If some token-level intent is incorrectly predicted, the other correct token-level prediction can still be useful for corresponding SF. For the final IC prediction, they use a voting mechanism to take into account the IC prediction on each token.  
 
\newcite{Chen2019BERTFJ} use a Transformer \cite{Vaswani2017AttentionIA} model for joint SF and IC by fine-tuning a pre-trained BERT \cite{Devlin2019BERTPO} model (Figure \ref{fig:sharing_gating_mechanism} \textit{Right}). The input is passed through several layers of transformer encoders and the hidden state outputs are  used to compute slot and intent labels. The hidden state $h^{\texttt{CLS}}$ is  used for IC\footnote{\texttt{[CLS]} is a special token in BERT input format that often used as the sentence representation.} while the rest of the hidden states at each time step $h_i$ serve SF.

\subsection{Slot-Intent Gate Mechanism} 

In addition to parameter and state sharing, a separate network with a \textit{slot gating mechanism} was  introduced by \newcite{goo2018slot} to model the interaction between SF and IC more explicitly (Figure \ref{fig:sharing_gating_mechanism} \textit{Middle}). In the encoder, a \textit{slot context vector} for each time step, $\boldsymbol{c}^{S}_{i}$, and a global intent context vector $\boldsymbol{c}^{I}$ are computed using an attention mechanism \cite{DBLP:journals/corr/BahdanauCB14}. The slot-gate $\boldsymbol{g}^{s}$ is computed as a function of $\boldsymbol{c}^{S}_{i}$ and $\boldsymbol{c}^{I}$, $\boldsymbol{g}^{s} = \sum v \cdot tanh(\boldsymbol{c}^{S}_{i} + \boldsymbol{W}\cdot \boldsymbol{c}^{I} )$. Then, $\boldsymbol{g}^{s}$ is used as a weight between $\boldsymbol{h}_i$ and $\boldsymbol{c}^{S}_i$ to compute $y^{slot}_{i}$ as follows: $y^{slot}_i=\texttt{softmax}(\boldsymbol{W}(\boldsymbol{h}_i + \boldsymbol{g}^{s}\cdot\boldsymbol{c}^{S}_i))$. Larger $\boldsymbol{g}^{s}$ indicates a stronger correlation between $c^{S}_{i}$ and $c^{I}$. 
% Their results show +1\% improvement over \citet{Liu2016AttentionBasedRN} for SF and IC on ATIS and SNIPS.  

 \newcite{Haihong2019ANB} propose a bi-directional model, SF-ID (SF-Intent Detection) network, sharing ideas with \newcite{goo2018slot}, with  two key differences. First, in addition to the slot-gated mechanism, they add an intent-gated mechanism as well. Second, they use an iterative mechanism between the SF and ID network, meaning that the gate vector from SF is injected into the ID network and vice versa. This  mechanism is repeated for an arbitrary number of iteration. Compared to \cite{goo2018slot}, the SF-ID network performs better both in SF and IC on ATIS and SNIPS. The work from \newcite{Li2018ASM} is also similar to \newcite{goo2018slot} with two differences. First, they use a self-attention mechanism \cite{Vaswani2017AttentionIA} to compute $c^{S}_{i}$. Secondly, they use a separate network to compute gate vector $g^{s}$, but the input of this network is the concatenation of $c^{S}_{i}$ and the intent embedding $v$, and $g^s$ is defined as $g^{s} = \texttt{tanh}(\boldsymbol{W}^{g}[c^i_{slot}, v^{intent}] + b^s)$. After that, $h_i$ is combined with $g^s$ through element-wise multiplication to compute $y^s_{i}$ as follows: $y^{slot}_i=\texttt{softmax}(\boldsymbol{W}^s(h_i \odot g^s) + b^s$). They report a +0.5\% improvement on SF over \newcite{Liu2016AttentionBasedRN}. A recent work by \newcite{Zhang2019AJL}, further improves the performance of the BERT based model by adding a gate mechanism \cite{Li2018ASM} to the BERT model. Table \ref{tab:joint_models} compares the performance of the joint models. 

\begin{table}[htb!]
\centering
\small
\begin{tabular}{llcccc}
\toprule
\multirow{3}{*}{\textbf{Method}} & \multirow{3}{*}{\textbf{Model}}& \multicolumn{2}{c}{\textbf{ATIS}} & \multicolumn{2}{c}{\textbf{SNIPS}} \\
% \cmidrule(lr){2-3} \cmidrule(lr){4-5}
 & & \textbf{Slot} & \textbf{Intent }& \textbf{Slot} & \textbf{Intent} \\
 & & \textbf{F1} & \textbf{Acc/Err}& \textbf{F1} & \textbf{Acc/Err} \\
 \midrule
% \textbf{Independent Model} &  &  &  &  \\
% CNN + Tri-CRF \cite{xu2013convolutional} & X & X   & - & - \\
% \midrule
\textbf{Parameter \& State Sharing}  & &  &  &  &  \\
 \newcite{xu2013convolutional} &CNN + Tri-CRF & 95.42 & -/5.91  & - & - \\
\newcite{guo2014joint} & Recursive NN & 93.96 & 95.40 & - & -\\
\newcite{Zhang2016AJM} & Joint Multi-Task,Bi-GRU & 95.49 & 98.10 & - & - \\
\newcite{Liu2016AttentionBasedRN} & Seq2Seq + Attention & 94.20 & 91.10 & 87.80 & 96.70 \\
\newcite{HakkaniTr2016MultiDomainJS} & Bi-LSTM& 94.30 & 92.60 & 87.30 & 96.90 \\
\newcite{Qin2019ASF}& Token-Level IC + Self-Attention & 95.90 & 96.90 & 94.20 & 98.00 \\

\newcite{Chen2019BERTFJ} & Transformer (BERT)& 96.10 & 97.50 & 97.00 & 98.60 \\
\midrule
\textbf{State Sharing} & &  &  &  &  \\
\newcite{Wang2018ABB} &Bi-model, BiLSTM  & 96.89 & 98.99 & - & - \\
\midrule
\textbf{Slot-Intent Gating} & &  &  &  &  \\
\newcite{goo2018slot}& Slot-Gated Full Attention \ & 94.80 & 93.60 & 88.80 & 97.70 \\
\newcite{Li2018ASM}& BiLSTM + Self-Attention & 96.52 & -/1.23 & -  & - \\
\newcite{Haihong2019ANB}& SF-ID Network  & 95.75 & 97.76 & 91.43 & 97.43 \\
\midrule
\textbf{Hybrid Param Sharing + Gating}&  &  &  &  &  \\
\newcite{Zhang2019AJL} & \textsc{BERT} + Intent-Gate  & 98.75 & 99.76 & 98.78 & 98.96 \\
\bottomrule
\end{tabular}
\caption{Performance comparison of joint models for SF and IC on ATIS and SNIPS-NLU.}
\label{tab:joint_models}
\end{table}

\textbf{Takeaways on joint SF and IC models:}
\begin{itemize}[noitemsep,nolistsep, leftmargin=*]
    \item The overall performance of joint models for SF and IC (Table \ref{tab:independent_models}) is competitive with independent models (Table \ref{tab:joint_models}). The advantage of joint models is that they have relatively less parameters than independent models, as both tasks are trained on a single model.   
    \item When computational power is not an issue, fine-tuning a pre-trained model such as BERT is the way to go for maximum SF and IC performance. Hybrid methods combining parameter and state sharing + intent gating yield the best performance \cite{Zhang2019AJL}. 
    \item For the non BERT-based model, using state sharing \cite{Wang2018ABB} is the best on ATIS. However, the disadvantage is that it is actually a bi-model and not a single model. 
    \item Similar to independent models, contextual information is crucial to performance. Adding a self-attention mechanism \cite{Qin2019ASF,Li2018ASM} to either parameter and state sharing or to slot-intent gating can boost performance even further.
    \item When sufficiently large in-domain training data is available, the SF and IC performance in ATIS and SNIPS is already saturated. Therefore, further research on this classic leaderboard chase is not worth it. We discuss more about that in Section 6. 
    \item Most of the work in joint models and also independent models (Section \textsection \ref{sec:independent_model}) reports F1 scores for slot filling performance. However, these scores do not reveal in which specific cases these models behave differently, contributing to overall performance. We leave further analysis on model performance as a potential future work.
\end{itemize}

\section{Scaling to New Domains}
\label{sec:transfer_learning}

So far, the models that we consider in Section \textsection \ref{sec:independent_model} and Section \textsection \ref{sec:joint} are  designed to be trained on a \textit{single domain} (e.g. banking, restaurant reservation) and require relatively \textit{large labeled data} to perform well. In practice, new intents and slots are regularly added to a system to support new tasks and domains,  requiring data and time intensive processes. Hence, methods to train models for new domains with limited or without labeled data are needed. We refer to this situation as the \textit{domain scaling} problem. 
\begin{figure*}[h!]
     \centering
     \includegraphics[scale=0.3]{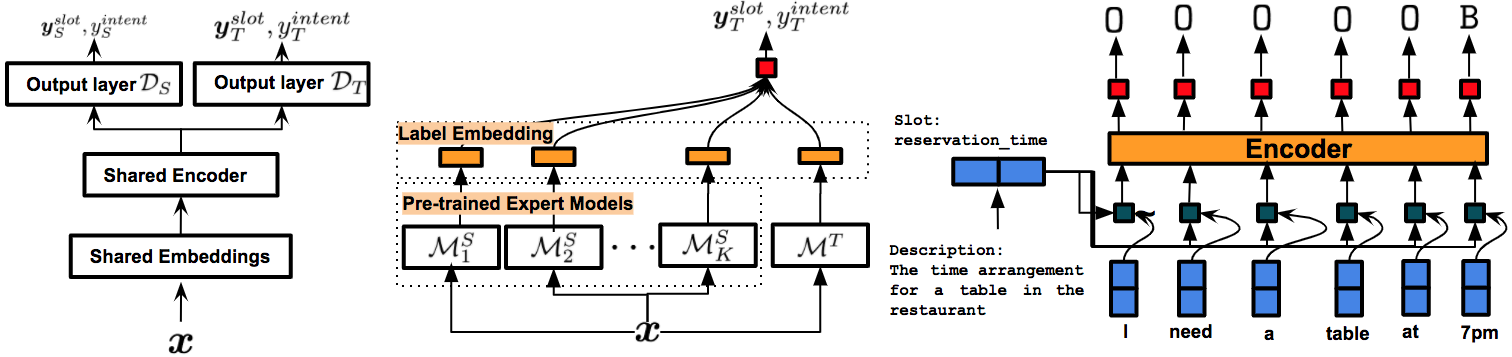}
    \caption{\textit{Left:} Data-driven approach \protect\cite{Jaech2016DomainAO,HakkaniTr2016MultiDomainJS}. \textit{Middle:} Model-Driven Approach with expert models \protect\cite{Kim2017DomainAW}. \textit{Right:} Zero-shot model \protect\cite{DBLP:conf/interspeech/BapnaTHH17}.}
    \label{fig:transfer_learning}
\end{figure*}
\subsection{Transfer Learning Models for SF and IC}
A common approach to deal with domain scaling is  transfer learning (TF).\footnote{We do not differentiate between \textit{domain adaptation} and transfer learning in this paper.}  In the TF setup we have $K$ source domains $\mathcal{D}^1_{S}, \mathcal{D}^{2}_{S}, \ldots, \mathcal{D}^{K}_{S}$ 
and a target domain $\mathcal{D}^{K+1}_{T}$, and we assume abundance of data in $\mathcal{D_S}$ and limited data in $\mathcal{D_T}$. Instead of 
training a target model $\mathcal{M}_T$ for $\mathcal{D}_T$ from scratch, TF aims to \textit{adapt} the learned model $\mathcal{M}_S$ from $\mathcal{D}_S$ to produce a model $\mathcal{M}_T$ trained on $\mathcal{D}_T$.  TF is typically applied with various parameter sharing and training mechanisms. For SF and IC  two approaches are proposed, namely \textit{data-driven} and \textit{model- driven}. As for data-driven techniques, typically we combine data from $\mathcal{D}_S$ and $\mathcal{D}_T$ and we partition the parameters in the model into parts that are \textit{task-specific} and parameters that are shared across tasks.  
Some studies \cite{Jaech2016DomainAO,HakkaniTr2016MultiDomainJS,Louvan2019LeveragingNT}  apply this technique using \textit{multi-task learning} (MTL) \cite{caruana1997multitask} and the models are trained simultaneously on $\mathcal{D}_S$ and $\mathcal{D}_T$ (Figure \ref{fig:transfer_learning} \textit{Left}). Results have shown that MTL is particularly effective relative to single-task learning (STL) when the data in $\mathcal{D_T}$ is scarce and the benefits over STL diminish as more data is available. Another technique that is typically used in  data-driven approaches is based on \textit{pre-train} and \textit{fine-tune} mechanisms. \newcite{DBLP:conf/naacl/GoyalMM18} train a joint model of SF and IC, $\mathcal{M}_S$, on large $\mathcal{D_S}$, then fine-tune $\mathcal{M}_S$ by replacing the output layer corresponding with the label space from $\mathcal{D}_T$ and train the model further on $\mathcal{D}_T$. \newcite{DBLP:conf/aaai/SiddhantGM19} also uses fine-tuning mechanism, but the main difference with \newcite{DBLP:conf/naacl/GoyalMM18} is they leverage large unlabeled data to learn contextual embedding, ELMo \cite{peters2018deep}, before fine-tuning on $\mathcal{D}_T$. 

As we need to train from scratch the whole model when adding a new domain,  data-driven approaches, especially MTL-based, need increasing training time as the number of domains grows. The alternative strategy, the model-driven approach, alleviates the problem by enabling model \textit{reusability}. Although different domains have different slot schemas,   slots such as \textit{date}, \textit{time} and \textit{location} can be shared. In model driven adaptation "expert" models (Figure \ref{fig:transfer_learning} \textit{Middle}) are first trained on these reusable slots \cite{Kim2017DomainAW,Jha2018BagOE} and the outputs of the expert models are used to guide the training of $\mathcal{M}_T$ for a new target domain. This way the training time of $\mathcal{M}_T$ is faster, as it is proportional to the $\mathcal{D}_T $ data size, instead of the larger data size of the whole $\mathcal{D}_S$ and $\mathcal{D}_T$. In this model-driven settings, \newcite{Kim2017DomainAW} do not treat each expert model on each $\mathcal{D}_S$ equally, instead they use attention mechanism to learn a weighted combinations from the feedback of the expert models. \newcite{Jha2018BagOE} use a similar model as \newcite{Kim2017DomainAW}, however they do not use attention mechanism. For training the expert models, instead of using all available  $\mathcal{D}_S$, they build a repository  consisting of common slots, such as \textit{date}, \textit{time}, \textit{location} slots. The assumption is that these slots are potentially reusable in many target domains. Upon training $\mathcal{M}_S$ on this reusable repository, the output of $\mathcal{M}_S$  is directly used to guide the training of $\mathcal{M}_T$.

\subsection{Zero-shot Models for SF and IC}
While data-driven and  model-driven approaches can share  knowledge learned on different domains, such models are still trained on a pre-defined set of labels, and can not handle \textit{unseen} labels, i.e. not mapped to the existing schema. For example, a model trained to recognize a \textsc{Destination} slot, can not be used directly to recognize the slot \textsc{Arrival\_Location} for a new domain, although both slots are semantically similar. For this reason, researchers  have recently been working on \textit{zero-shot} models,  trained on \textit{label representations} that leverage  natural language \textit{descriptions} of the slots \cite{DBLP:conf/interspeech/BapnaTHH17,Lee2019ZeroShotAT}. Assuming that accurate slot descriptions are provided, slots with \textit{different} names  although semantically similar would have similar description as well. Thus, having trained a model for the \textsc{Destination} slot with its descriptions,  it is now possible to recognize the slot \textsc{Arrival\_Location} without training on it, but  only supplying the corresponding slot description.  

In addition to slot description, other zero-shot approaches  explore the use of slot value examples \cite{shah-etal-2019-robust,DBLP:conf/sigdial/GueriniMBM18}. \newcite{shah-etal-2019-robust} showing that a  combination of a small number of slot values examples  with a slot description performs better than \cite{DBLP:conf/interspeech/BapnaTHH17,Lee2019ZeroShotAT} on the SNIPS dataset. Zero-shot models are typically trained on a per-slot basis (Figure \ref{fig:transfer_learning} \textit{Right}), meaning that if we have $N$ slots, then the model will output $N$ predictions, therefore, a merging mechanism is needed in case there are prediction overlaps. In order to alleviate the problem of having multiple predictions, \newcite{DBLP:conf/acl/LiuWXF20} propose a \textit{coarse-to-fine} approach, in which the model learns the slot entity pattern (coarsely) to identify a particular token is an entity or not. After that, the model performs a single prediction of the slot type (fine) based on the similarity between the feature representation and the slot description. 

\paragraph{Takeaways on scaling to new domains:}
\begin{itemize}[noitemsep,nolistsep, leftmargin=*]
    \item Both data driven methods, MTL and pre-train fine tuning,  improve performance when data in $\mathcal{D}_T$ is limited. Both are also flexible, as virtually many tasks from different domains can be plugged into these methods. As the number of domains grow, pre-train and fine tuning is more desirable than MTL. However, fine tuning is more prone to the \textit{forgetting} problem \cite{He2019AnalyzingTF} compared to MTL.
    \item When the number of domain, $K$, is massive, the pre-train fine tuning approach and  model driven approaches, such as expert based adaptation, are preferable with respect of training time.  
    \item When there exists $K$ existing domains and no annotation is available in  $\mathcal{D}_T$, the choice is zero-shot approaches with the expense of providing meta-information such as slot and intent descriptions.
    \item As typically zero-shot models perform prediction on a \textit{per-slot} basis,  potential disadvantages are model accuracy when there is a prediction overlap and the model can also be computationally inefficient when dealing with many slots.
\end{itemize}
\section{State of the Art and Beyond}
\label{sec:discussion}

Based on the results in Table \ref{tab:independent_models} and \ref{tab:joint_models}, it is evident that  neural models have achieved outstanding performance on ATIS and SNIPS, showing that it is relatively easy for neural models to capture  patterns that recognize slots and intents. ATIS, in particular, is already overused for SF and IC evaluations and recent analysis \cite{Bchet2018IsAT,Niu2019RationallyRA} have  shown that the dataset is relatively simple  and the room for performance improvement is tiny. 
%Thus, although many innovations of neural models have been proposed, it is hard to see clear improvements as the performance gap between models are small. 
A similar trend in performance can be noted for other datasets, such as SNIPS, and it is likely that performance improvement can be quickly saturated. However, it does not mean these models have solved SF and IC, or NLU problems in general,  rather that the model has merely solved the datasets. 
Nevertheless, there are still a number of issues in  SF and IC that  need  further investigation:
\paragraph{Portable and Data Efficient Models.} Instead of  evaluating  models with the typical \textit{leaderboard} setup with fixed (train/dev/test) splits on a specific target domain, it would be also important to test  models in different scenarios, so that  different aspects of the model can be captured. For example, as neural models are data hungry, more work is still needed on transfer learning scenarios, where evaluation  is carried out with \textit{less} or \textit{without} labeled data (\textit{zero-shot}) for a particular target domain. In addition, most models for SF and IC are evaluated on English, which means that more effort is still needed to make models that work well for other languages. Some recent works have started exploring zero-shot cross lingual methods \cite{DBLP:conf/ijcai/QinN0C20,DBLP:conf/aaai/LiuWLXF20,DBLP:conf/emnlp/LiuSXWXMF19} and also few-shot scenarios \cite{DBLP:conf/acl/HouCLZLLL20} and the room for improvement for these scenarios is still large. In short, designing a \textit{data efficient} model that is \textit{portable} across domains and languages is still a challenging problem for the coming future.
    \paragraph{Leveraging unlabeled data from live traffic.} In real situations, personal digital assistants such as Google Home, Apple Siri and Amazon Alexa, receive live traffic data from real users. This large amount of unlabeled data from live traffic is a potential data source for model training, in addition to  in-house annotated data.  Unlabeled live data are likely different from in-house data, as they can contain more diverse utterances and also noisy and irrelevant utterances. In this situation, existing methods to tap on unlabeled data, such as semi-supervised learning, still face unique challenges to  handle live data. %Thus, methods to effectively capitalize the live traffic data is still very much needed. 
    It is worth to note that a bottleneck in this direction is that working on live data in academic settings is not trivial. Some recent works explore this line of research by applying semi-supervised learning \cite{DBLP:conf/asru/ChoXLKC19} and also data selection \cite{DBLP:conf/emnlp/DoG19} mechanism.
    \paragraph{Generative Models.} Most of the proposed models are \textit{discriminative}, among the few works  carried out for \textit{generative models} for SF and IC, \cite{Raymond2007GenerativeAD,Yogatama2017GenerativeAD} have shown that a generative model is relatively better than a discriminative model in a situation where data is \textit{scarce}. One possible direction for generative models is to apply data augmentation to automatically create additional training data \cite{DBLP:conf/aaai/YooSL19,DBLP:conf/emnlp/ZhaoZY19,DBLP:conf/coling/HouLCL18,DBLP:conf/emnlp/KurataXZY16,DBLP:journals/corr/abs-2004-13952,DBLP:conf/naacl/KimRK19}. The main challenge for data augmentation is to generate diverse and fluent synthetic utterance, which \textit{preserve} the semantics of the original utterance. 
    
    \paragraph{Evaluation of SF and IC on more complex dataset.} Existing neural approaches typically evaluated on \textit{single-intent} utterance, however in a real-world scenario users may indicate \textit{multiple-intent} in an utterance e.g. "\textit{Show me all flights from Atlanta to London and get the cost}" \cite{Gangadharaiah2019JointMI} or even expressing multiple sentences in one single turn. While most datasets for slot filling and intent classification are \textit{single-turn} utterance, there are some recent multi-turn datasets that provide slot annotation on the token-level, namely the \textsc{Restaurant-8K}, TaskMaster-1 and 2 \cite{48484}, and Frame \cite{DBLP:conf/sigdial/AsriSSZHFMS17} datasets. The subset of Schema Guided Dialogue (SGD) dataset \cite{DBLP:conf/aaai/RastogiZSGK20} used in DTSC-8 is also annotated with slots in the token-level and covers 16 domains. In addition to that, the TOP dataset \cite{DBLP:conf/emnlp/GuptaSMKL18} introduces datasets annotated with \textit{hierarchical} representation and MTOP dataset \cite{DBLP:journals/corr/abs-2008-09335} provides both flat and hierarchical representation on 6 languages across 11 domains.

\section{Conclusion}
\label{sec:conclusion}
We have surveyed recent neural-based models applied to SF and IC in the context of task-oriented dialogue systems. We examined three approaches, i.e.  \textit{independent}, \textit{joint}, and \textit{transfer learning based} models. Joint models  exploiting the relation between SF and IC simultaneously shown relatively better performance than independent models. Empirical results have shown that most joint  models nearly "solve"  widely used datasets,  ATIS and SNIPS, given \textit{sufficient in-domain training data}. Nevertheless, there are still several challenges related to SF and IC, especially  improving the scalability of the model to new domains and languages when limited labeled data are available.

\bibliographystyle{coling}
\bibliography{coling2020}

\section*{Appendix A. SF and IC Dataset}
\label{appendix:dataset}
\begin{table}[!htb]
\begin{center}
\begin{tabular}{llrrr}
\toprule

\bf Dataset  & \bf Language & \bf \# intent & \bf \# slot  & \bf \# sentences train/dev/test\\ \midrule
 ATIS & English & 18 & 83 & 4,478 / 500 / 893\\
 \midrule
 MEDIA & French & - & 68 & 12,908/1,259/3,005 \\
\midrule 
SNIPS-NLU & English  & 7 & 39 &  13,084 / 700 / 700\\
          & Italian  & 7 & 39 &  5,742 /700 / 700\\
\midrule
 Facebook & English  & 12 & 11 & 30,521 / 4,181 / 8,621 \\
 Multilingual         & Thai  & 12 & 11 & 3,617 / 1,983 / 3,043 \\
          & Spanish  & 12 & 11 & 2,156 / 1,235 / 1,692\\
\midrule
 MIT Restaurant & English  & - & 8 & 6,128 / 1,532 / 1,521  \\
 \midrule
 MIT Movie & English  & - & 12 & 7,820 / 1,955 / 2,443 \\

\bottomrule
\end{tabular}
\end{center}
\caption{\label{dataset-table} Single-turn datasets statistics. }
\end{table}

\end{document}